\documentclass{article}

    \PassOptionsToPackage{numbers, compress}{natbib}

\usepackage[final]{neurips_2021_ml4ps}

\input{macros}

\usepackage[utf8]{inputenc} 
\usepackage[T1]{fontenc}    
\usepackage{hyperref}       
\usepackage{url}            
\usepackage{booktabs}       
\usepackage{amsfonts}       
\usepackage{nicefrac}       
\usepackage{microtype}      
\usepackage{xcolor}         

\title{Multiway Ensemble Kalman Filter}

\author{%
  Yu Wang \\
  University of Michigan\\
  \texttt{wayneyw@umich.edu} \\
   \And
   Alfred Hero \\
   University of Michigan \\
   \texttt{hero@umich.edu} \\
}

\begin{document}

\maketitle

\begin{abstract}
In this work, we study the emergence of sparsity and multiway structures in second-order statistical characterizations of dynamical processes governed by partial differential equations (PDEs). We consider several state-of-the-art multiway covariance and inverse covariance (precision) matrix estimators and examine their pros and cons in terms of accuracy and interpretability in the context of physics-driven forecasting when incorporated into the ensemble Kalman filter (EnKF). In particular, we show that multiway data generated from the Poisson and the convection-diffusion types of PDEs can be accurately tracked via EnKF when integrated with appropriate covariance and precision matrix estimators.
\end{abstract}

\section{Introduction}\label{sec:intro}
There has recently been a resurgence of interest in integrating machine learning with physics-based modeling. Much of the recent work has focused on black-box models such as deep neural networks~\citep{takeishi2017learning, long2018pde, NEURIPS2018_e2ad76f2, vlachas2018data, reichstein2019deep, wang2020towards}. However, seeking shallower models that capture mechanism  in a physically  interpretable manner has been a recurring theme in both machine learning and physics~\citep{weinan2020integrating}. In this paper, we introduce a high-dimensional statistical approach that naturally integrates physics and machine learning through Kronecker-structured Gaussian graphical models. The learned representation can then be incorporated into a high dimensional predictive model using the ensemble Kalman filtering framework.

{\bf Multiway covariance/precision models.} High-dimensional multiway/tensor-variate data arise naturally in physical sciences. For example, weather satellites measure spatio-temporal climate variables such as temperature, wind velocity, sea level, pressure, etc. Due to the non-homogeneous nature of these data, second-order information that encodes (conditional) dependency structure within the data is of interest. Assuming the data are drawn from a tensor normal distribution, a straightforward way to estimate this structure is to vectorize the tensor and estimate the underlying Gaussian graphical model associated with the vector. However, such an approach ignores the tensor structure and requires estimating a rather high dimensional precision matrix, often with insufficient sample size. In many scientific applications the sample size can be as small as one when only a single tensor-valued measurement is available.

To address sample complexity challenges in learning second-order representations for multiway (tensor) data, sparsity is often imposed on the covariance $\mat{\Sigma}$ or the inverse covariance $\mat{\Omega}$. Such approaches include the sparse Kronecker product (KP) or Kronecker sum (KS) decomposition of $\mat{\Sigma}$ or $\mat{\Omega}$. Statistical models and corresponding learning algorithms can be derived using generative models or matrix approximations. The former include: KGlasso/Tlasso~\citep{tsiligkaridis2013convergence,lyu2019tensor} for estimating $\mat{\Omega}=\mat{A} \otimes \mat{B}$, using a representation $\mat{A}\mat{X}\mat{B} = \mat{Z}$ for data $\mat{X}$ when $\mat{Z}$ is white noise. Another generative model is SyGlasso/SG-PALM~\citep{wang2020sylvester,wang2021sg} that models the precision matrix as $\mat{\Omega}=(\mat{A} \oplus \mat{B})^2$, which corresponds to assuming the data $\mat{X}$ obeys a Sylvester equation $\mat{X} \mat{A} + \mat{B} \mat{X} = \mat{Z}$. Matrix approximation methods include: KPCA~\citep{tsiligkaridis2013covariance,greenewald2015robust} that approximates the covariance matrix as $\mat{\Sigma}=\sum_{i=1}^r \mat{A}_i \otimes \mat{B}_i$. Another matrix approximation method is the TeraLasso~\citep{greenewald2019tensor} that models the precision matrix as $\mat{\Omega}=\mat{A} \oplus \mat{B}$. TeraLasso is equivalent to approximation of the conditional dependency graph (encoded by the precision matrix) with a Cartesian product of smaller graphs~\footnote{Note that Tlasso, TeraLasso, Syglasso/SG-PALM are generalizable to precision matrices of the form $\bigotimes_{k=1}^K \mat{A}_k$, $\bigoplus_{k=1}^K \mat{A}_k$, and $(\bigoplus_{k=1}^K \mat{A}_k)^2$, respectively, for $K \geq 2$.}. 

{\bf Multiway second-order characterization of dynamic processes.} Physical systems often exhibits sparsity and low-rank structures in their covariance or inverse covariance matrix. This is due to the fact that many physical systems are governed by differential equations, which are characterized by sparse differential operators. For instance, \citet{wang2021sg} showed that for multivariate data generated by the Poisson equation, the discretized data has an inverse covariance matrix equal to a squared Kronecker sum of smaller sparse matrices. In related work, \citet{lindgren2011explicit} elucidated a link between certain classes of Gaussian Fields (GF) and Gaussian Markov Random Fields (GMRF) via stochastic partial differential equations, and showed that efficient learning algorithms can be developed using the fact that GMRFs have sparse precision matrices.

For multiway/tensor-variate Gaussian data, the aforementioned multiway (inverse) covariance estimators have been shown to be statistically consistent in high-dimensional regimes when sample sizes ($N$) are much less than the dimensionality ($d$) of the covariates. An important question is whether these Kronecker structures can be integrated into the
Kalman filter for tracking the states of a physical system that generates multiway data.

\section{Numerical experiments: ensemble Kalman filtering}\label{sec:numerics}
The Kalman filter is a well-known technique to track the states of a linear system over time, and many variants have been proposed to deal with non-linear systems, such as the extended and ensemble Kalman filters. The ensemble Kalman filter (EnKF) is particularly effective when the dynamical system is complicated and its gradient is infeasible to calculate, which is often the case in physical systems~\citep{evensen1994sequential,burgers1998analysis}. However, such systems are often high-dimensional and the EnKF operates in the regime where the number of ensemble members, $N$, is much less than the size of the state, $d$, suggesting that sparse inverse covariance models will be especially attractive. \citet{hou2021penalized} introduced a sparsity-penalized EnKF, which uses an estimator of the forecasting covariance whose inverse is sparsity regularized. Here we propose incorporating the multiway covariance / inverse covariance models discussed in Section~\ref{sec:intro} into the EnKF of \citet{hou2021penalized}.

For motivation, we consider the Poisson equation, an elliptical PDE that governs many physical processes including electromagnetic induction, heat transfer, and convection~\citep{chandrasekhar1943stochastic}. On a rectangular region $\Omega=(0,d_1)\times(0,d_2)$ in the 2D Cartesian plane, the Poisson equation with homogeneous Dirichlet boundary condition is expressed as
\begin{equation}
    \begin{aligned}
        \mathcal{D}u = (\partial^2_x + \partial^2_y)u &= f \quad \text{in } \Omega, \\
        u &= 0 \quad \text{on } \partial\Omega
        \label{eq:Poisson}
    \end{aligned}
\end{equation}
where $f: \Omega \to \bbR$ is the given source function and $u: \Omega \to \bbR$ is the unknown. Using the finite difference method with a square mesh grid with unit spacing, the unknown and the source can be expressed as $d_1$-by-$d_2$ matrices, $\bU$ and $\bF$, respectively, that are related to each other via
\begin{align}\label{eqn:discrete-poisson}
    U_{i+1,j} + U_{i-1,j} + U_{i,j+1} + U_{i,j-1} - 4 U_{i,j} = F_{i,j}
\end{align}
for any interior grid point $(i,j)$. Defining $n$-by-$n$ square matrix
\begin{equation*}
\mat{A}_n = 
    \begin{bmatrix}
    2   &   -1  &       &   \\
    -1  &   2   & \ddots&   \\
        & \ddots& \ddots& -1\\
        &       &   -1  & 2
    \end{bmatrix},
\end{equation*}
the relation~\eqref{eqn:discrete-poisson} can be expressed as the (vectorized) Sylvester equation with $K=2$:
\begin{equation}
    (\bA_{d_1} \oplus \bA_{d_2})\bu = \bff,
    \label{eq:poisson_discrete}
\end{equation}
where $\bu = \vecto(\bU)$, $\bff = \vecto(\bF)$. Note that $\mat{A}$ is tridiagonal. In the case where $\mat{f}$ is white noise with variance $\sigma^2$, the inverse covariance matrix of $\bu$ has the form $\cov^{-1}(\bu)=\sigma^{-2}(\bA_{d_1} \oplus \bA_{d_2})^T(\bA_{d_1} \oplus \bA_{d_2})$ and hence sparse.

Here, we discuss two ways to extend the spatial Poisson equation described above to incorporate temporal dynamics, and illustrate how multiway (inverse) covariance models can be used to track spatio-temporal systems. The first extension, which we call the Poisson-AR(1) process, imposes an autoregressive temporal model of order 1 on the source function $f$ in the Poisson equation (\ref{eq:Poisson}). Specifically, we say a sequence of discretized spatial observations $\{\bU^k \in \bbR^{d_1\times d_2}\}_k$ indexed by time step $k=1,\cdots,T$ is from a Poisson-AR(1) process if
\begin{align}
    &(\mat{A}_{d_1} \oplus \mat{A}_{d_2}) \vecto(\mat{U}^k) = \vecto(\mat{Z}^k), \\
    &\vecto(\mat{Z}^k) = a \vecto(\mat{Z}^{k-1}) + \vecto(\mat{W}^k),\quad |a|<1, \label{eq:ar1}
\end{align}
where $\mat{Z}^0 \sim \mathcal{N}(\mat{0}, \sigma^2_z \mat{I})$ and $\{\mat{W}^k \in \bbR^{d_1\times d_2}\}_k$ is spatiotemporal white noise, i.e., $W_{i,j}^k \sim \mathcal{N}(0, \sigma^2_w)$, i.i.d.

The second time-varying extension of the Poisson PDE model (\ref{eq:Poisson}) is the convection-diffusion process~\cite{chandrasekhar1943stochastic}
\begin{equation}\label{eqn:convec-diff}
    \frac{\partial u}{\partial t} = \theta \sum_{i=1}^2 \frac{\partial^2 u}{\partial x_i^2} - \epsilon \sum_{i=1}^2 \frac{\partial u}{\partial x_i}.
\end{equation}
Here, $\theta > 0$ is the diffusivity; and $\epsilon \in \Reals$ is the convection velocity of the quantity along each coordinate. Note that for simplicity of discussion here, we assume these coefficients do not change with space and time (see, \citet{stocker2011introduction}, for example, for a detailed discussion). These equations are closely related to the Navier-Stokes equation commonly used in stochastic modeling for weather and climate prediction~\citep{chandrasekhar1943stochastic,stocker2011introduction}. Coupled with Maxwell's equations, these equations can be used to model magneto-hydrodynamics~\citep{roberts2006slow}, which characterize solar activities including flares. 

A solution of Equation~\eqref{eqn:convec-diff} can be approximated similarly as in the Poisson equation case, through a finite difference approach. Denote the discrete spatial samples of $u(\mat{x},t)$ at time $t_k$ as a matrix $\mat{U}^k\in\bbR^{d_1 \times d_2}$. We obtain a discretized update propagating $u(\mat{x},t)$ in space and time, which locally satisfies
\begin{equation}\label{eqn:discrete-convection-diffusion}
\begin{aligned}
    \frac{U_{i,j}^k - U_{i,j}^{k-1}}{\Delta t} = &\ \theta \left(\frac{U_{i+1,j}^k + U_{i-1,j}^k + U_{i,j+1}^k + U_{i,j-1}^k - 4U_{i,j}^k}{h^2}\right) \\
    &- \epsilon \left(\frac{U_{i+1,j}^k - U_{i-1,j}^k + U_{i,j+1}^k - U_{i,j-1}^k}{2h}\right),
\end{aligned}
\end{equation}
where $\Delta t = t_{k+1} - t_{k}$ is the time step and $h$ is the mesh step (spatial grid spacing). For $\Delta t = 1$ and $h = 1$, $\mat{U}^k$ can be shown to obey the Sylvester matrix update equation~\citep{thomas2013numerical} $\mat{A}_{\theta,\epsilon}\mat{U}^k + \mat{U}^k\mat{B}_{\theta,\epsilon}^T = \mat{U}^{k-1}$,
or equivalently,
\begin{align}
    (\mat{B}_{\theta,\epsilon} \oplus \mat{A}_{\theta,\epsilon})\vecto({\mat{U}}^k) =\vecto({\mat{U}}^{k-1}),
    \label{eq:CD_discrete_update}
\end{align}
where $\mat{A}_{\theta,\epsilon}\in\bbR^{d_1\times d_1}$ and $\mat{B}_{\theta,\epsilon}\in\bbR^{d_2\times d_2}$ are symmetric tridiagonal matrices whose entries depend on $\theta, \epsilon$, and $h$ \cite{grasedyck2004existence}. Note that, although $\{\mat{U}^k\}_k$ updates according to a sparse Kronecker sum, the inverse covariance of the marginal distribution that smooths over time is not sparse. 

We turn Equation (\ref{eq:CD_discrete_update}) into a state space model for the observations $\mat{X}$ by adding i.i.d. state noise $\mat{W}^k \sim \mathcal N(0, \sigma^2_w \mat{I})$, i.i.d. observation noise $\mat{V}^k\sim \mathcal N(0, \sigma^2_v  \mat{I})$  and a state measurement matrix $\mat{H}$:    
\begin{align}
    \vecto(\mat{X}^k) &= \mat{H} \vecto(\mat{U}^k) + \vecto(\mat{V}^k),
    \\
    \vecto(\mat{U}^k) &= (\mat{B}_{\theta,\epsilon} \oplus \mat{A}_{\theta,\epsilon})^{-1} \vecto(\mat{U}^{k-1}) + \vecto(\mat{W}^k). 
\end{align}
We model the observed process $\mat{X}$ as an incomplete noisy version of the convection-diffusion state $\mat{U}_t$ 
obeying the Sylvester matrix update equation above
discretization the convection-diffusion becomes  assume a linear Gaussian 
state-space model for the observed process $\mat{X}_t$ governed by convection-diffusion dynamics:
\begin{equation*}
    \begin{aligned}
        & \mat{A}_{\theta,\epsilon}\mat{U}_t + \mat{U}_t\mat{A}_{\theta,\epsilon} = \mat{U}_{t-1}, \\
        & \mat{X}_t = \mat{U}_t + \mat{V}_t,
    \end{aligned}
\end{equation*}
where $\mat{V}_t \sim \mathcal{N}(\mat{0},\sigma^2\mat{I})$ is an i.i.d. Gaussian white noise. Note that in general one might not fully observe the state, leading to a partially observed measurement process
\begin{equation*}
    \vecto(\mat{X}_t) = \mat{H} \vecto(\mat{U}_t) + \vecto(\mat{V}_t),
\end{equation*}
where $\mat{H}$ is a measurement matrix that can incorporate effects such as unobserved, masked, or superposed states. It is also possible to take into account the error in the dynamic process, i.e.,
\begin{equation*}
    \vecto(\mat{U}_{t}) = (\mat{A}_{\theta,\epsilon} \oplus \mat{A}_{\theta,\epsilon})^{-1} \vecto(\mat{U}_{t-1}) + \vecto(\mat{W}_t,) 
\end{equation*}
where $\vecto(\mat{W}_t)$ is assumed to be white noise. 
Under the case of perfect observation ($\mat{H}$ is the identity matrix and $\mat{W}_t=\mat{0}$), 
Note that although the state variable evolves via a Sylvester equation, similar to the Poisson equation case, the state (inverse) covariance matrix at time step $t_k$ admits different structures. Specifically, the state precision matrix $\mat\Omega^k=\cov^{-1}(\vecto({\mat{U}^k})) \in \Reals^{d_1d_2 \times d_1d_2}$ evolves as $\mat\Omega^k = (\mat{B}_{\theta,\epsilon} \oplus \mat{A}_{\theta,\epsilon}) \mat\Omega^{k-1} (\mat{B}_{\theta,\epsilon} \oplus \mat{A}_{\theta,\epsilon}) + \sigma^{-2}_w\mat{I}$ (see \citet{katzfuss2016understanding}, for example). This matrix is not necessarily sparse for finite $k$ but, assuming that the eigenvalues of the matrix $\mat{B}_{\theta,\epsilon} \oplus \mat{A}_{\theta,\epsilon}$ are in $(-1,1)$, the limiting precision matrix  $\mat\Omega^{\infty}=\lim_{k\rightarrow\infty} \mat\Omega^k$ is $\mat\Omega^{\infty} = (\mat{B}_{\theta,\epsilon} \oplus \mat{A}_{\theta,\epsilon}) \mat\Omega^{\infty} (\mat{B}_{\theta,\epsilon} \oplus \mat{A}_{\theta,\epsilon}) + \sigma^{-2}\mat{I}$. The $\mat\Omega^{\infty}$ matrix is sparse because $\mat{A}_{\theta,\epsilon}$ and $\mat{B}_{\theta,\epsilon}$ are both tridiagonal.

\begin{figure}[!tbh]
    \centering
    \includegraphics[width=0.5\textwidth]{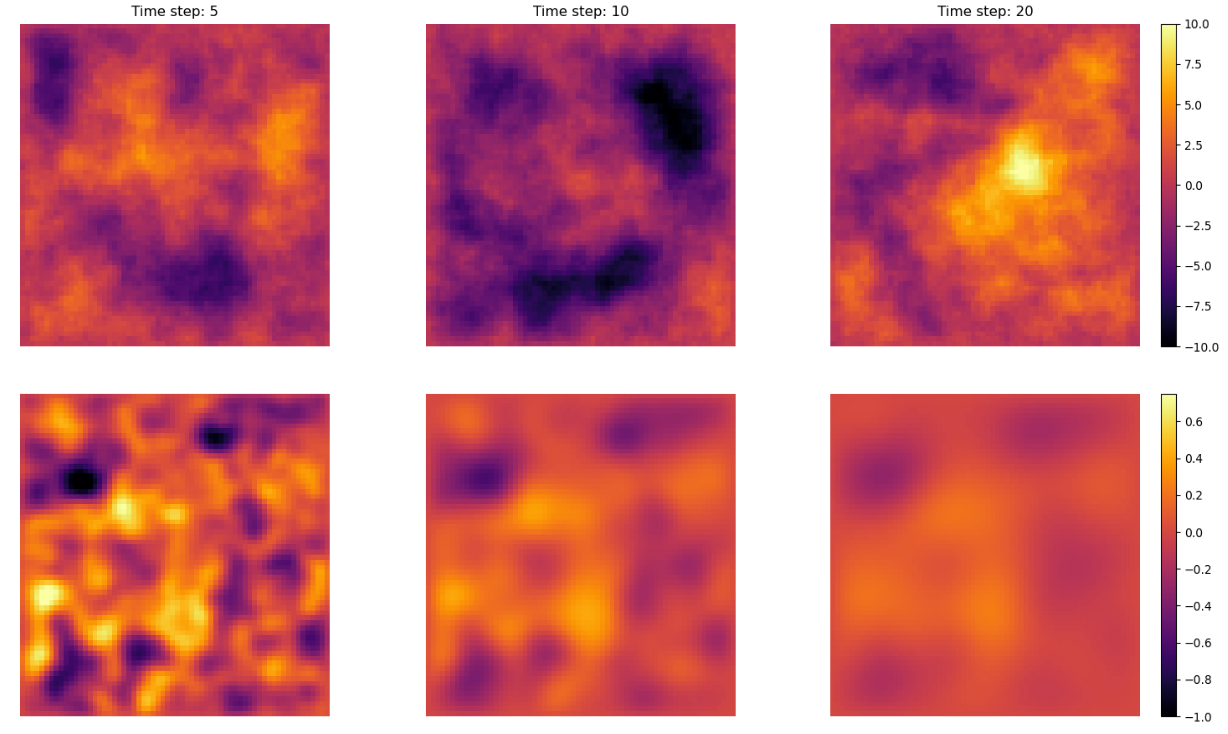}
    \caption{2D convection-diffusion (bottom) and Poisson (top) state variables at three different time stamps. Note that there is temporal correlation exists in the convection-diffusion states while the Poisson states are temporally independent.}
    \label{fig:enkf_states}
\end{figure}

\begin{figure}[!htb] 
\centering
\begin{tabular}{@{}cc@{}}
\qquad \qquad  Noise-free states & \quad  Noisy states\\
\rotatebox{90}{\quad Convection-Diffusion} \qquad 
\includegraphics[width=0.35\linewidth]{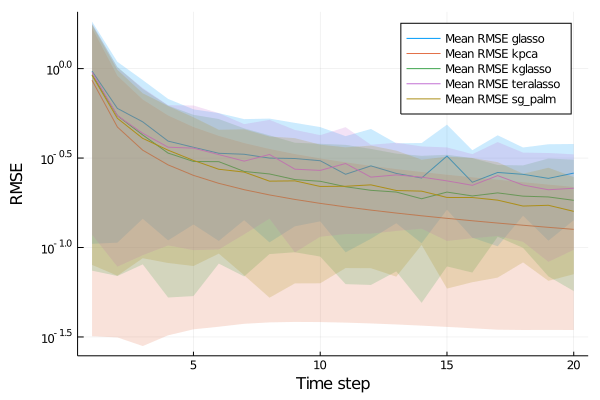}
&
\includegraphics[width=0.35\linewidth]{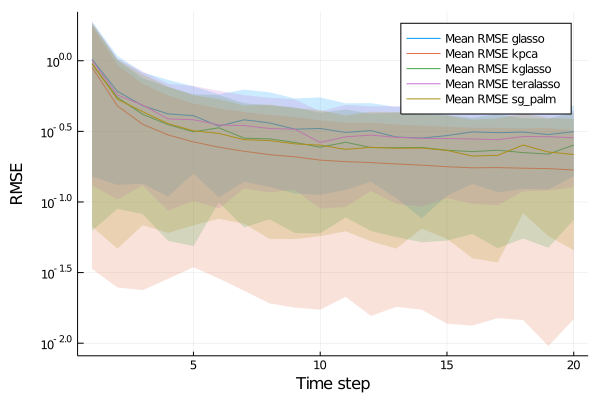}\\
\rotatebox{90}{\qquad \quad Poisson} \qquad  
\includegraphics[width=0.35\linewidth]{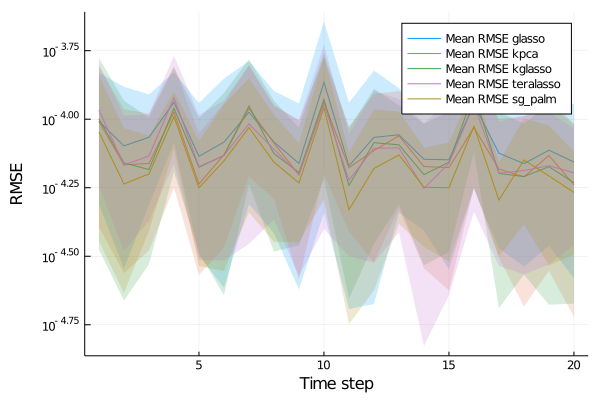}
& 
\includegraphics[width=0.35\linewidth]{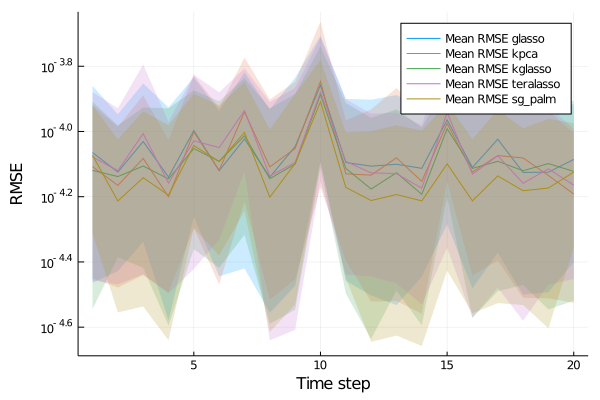}
\end{tabular}
\caption{RMSEs of the estimated states via EnKF over $20$ time steps using different (inverse) covariance estimators. RMSEs over all ensemble members are shown here with the mean highlighted using solid lines. Here, each state is of dimension $64 \times 64$ and is generated via either a convection-diffusion (top row) or Poisson equation (bottom row). The best performers in terms of mean RMSE over all ensemble members are KPCA for convection-diffusion and SG-PALM for Poisson.}
\label{fig:nrmse_enkf}
\end{figure}

\begin{figure}[!tbh]
    \centering   
    \includegraphics[width=0.7\textwidth]{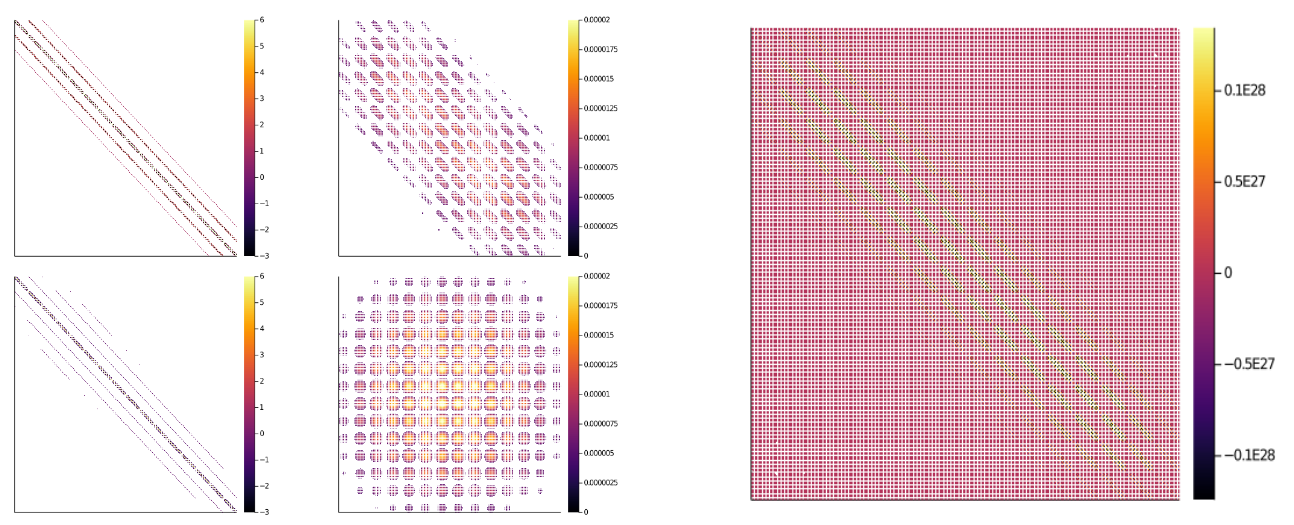}
    \caption{Covariance/precision structures for Poisson and Convection-Diffusion dynamics and their estimates. Here, white/blank entries indicate zeros in the (inverse) covariance matrix. For Poisson dynamics the Sylvester graphical model matches the true structure of the precision matrix. For convection-diffusion dynamics the covariance instead of the precision matrix is structured and sparse.}
    \label{fig:inv_cov_struct}
\end{figure}

To illustrate, we consider a 2D spatio-temporal process of dimension $64 \times 64$ where only half of the entries are observed, which leads to a measurement matrix $\mat{H} \in \{0,1\}^{2048 \times 4069}$. We generated the true states and the corresponding observations according to Convection-Diffusion and Poisson dynamics for $T = 20$ time steps. Several realizations of the true state variables are shown in Figure~\ref{fig:enkf_states}. At each time step, we generated an ensemble of size $N = 25$ and estimated the state covariance / inverse covariance using several sparse (multiway) inverse covariance estimation methods, including Glasso~\citep{friedman2008sparse}, KPCA~\citep{greenewald2015robust}, KGlasso~\citep{tsiligkaridis2013convergence}, TeraLasso~\citep{greenewald2019tensor}, SG-PALM~\citep{wang2021sg}. Figure~\ref{fig:nrmse_enkf} shows evolution of the computed root mean squared errors (RMSEs) for the estimated states across all ensemble members. In Figure~\ref{fig:inv_cov_struct} we show the true and estimated (inverse) covariance matrices obtained for the last time step. The Poisson process is time-invariant, and at each time step the EnKF involves estimation of a sparse Kronecker sum squared inverse covariance matrix. Hence, the SG-PALM method operates under the correct model assumption in this situation. On the other hand, the inverse covariance structure under the convection-diffusion dynamics model is dense due to the smoothing effect of Kalman filtering and the nature of the temporal dynamics. But, its steady-state covariance has low-dimensional structures. The KPCA in this case was able to approximate this structure, as illustrated in Figure~\ref{fig:inv_cov_struct}.  

\textit{Remark.} The proposed multiway EnKF is able to track systems governed by elliptic (e.g., Poisson) parabolic, and hyperbolic (e.g., convection-diffusion) PDEs. There are indeed other important PDEs / dynamical systems that cannot be modelled by these types of equations. Furthermore, the Sylvester matrix equations arise when the finite-difference discretization is performed on a rectangular grid. The relations~\eqref{eqn:discrete-poisson} and \eqref{eqn:discrete-convection-diffusion} might not hold for finite-difference on, for example, spherical coordinates.

\section{Future work}\label{sec:conclude}
\paragraph{Applications.} Spatiotemporal PDEs are prominent techniques for modeling real-world physical systems. One such system arises in space physics, where solar flares and coronal mass ejections are associated with rapid changes in filed connectivity and are powered by partial dissipation of electrical currents in the solar atmosphere~\citep{schrijver2008nonlinear}. The nonlinear force-free filed model is often used to describe the solar coronal magnetic field~\citep{derosa2015influence,wheatland2013state} and can be derived from the convection-diffusion process described in this work. Additionally, global maps of the solar photospheric magnetic flux are fundamental drivers for simulations of the corona and solar wind. However, observations of the solar photosphere are only made intermittently over approximately half of the solar surface. \citet{hickmann2015data} introduced the Air Force Data Assimilative Photospheric Flux Transport model that uses localized ensemble Kalman filtering to adjust a set of photospheric simulations to agree with the available observations. In future work, we plan to incorporate our proposed multiway EnKF framework for tracking these solar physical systems. 

\paragraph{Interpretability of factorization-based multiway model.} While the Kronecker products expansion used in KPCA captures dense structures in the covariance matrix of data generated from more complex spatio-temporal physical processes, it lacks physical interpretability. In contrast to the case of Sylvester graphical model and Poisson processes, it is not obvious whether the sum of Kronecker products structure corresponds to any true physical models. Recent work in quantum informatics~\citep{chu2021nonlinear} has demonstrated a link between estimation of the density matrix for entangled quantum states and the structured tensor approximation via $\sum_{i=1}^r \mat{A}_i \otimes \mat{B}_i$. Further characterizing these connections and extending them to study its connections with certain classes of discretized PDEs would be an interesting future direction.

\clearpage

\begin{ack}
The authors thank Zeyu Sun at the University of Michigan for many helpful discussions of potential applications of the work on solar physics. The work is partially supported by NASA DRIVE Science Center grant 80NSSC20K0600 and by the National Nuclear Security Administration within the US Department of Energy under grant DE-NA0003921.
\end{ack}

\bibliographystyle{plainnat}
\bibliography{neurips_2021_ml4ps}

\section*{Checklist}


\begin{enumerate}

\item For all authors...
\begin{enumerate}
  \item Do the main claims made in the abstract and introduction accurately reflect the paper's contributions and scope?
    \answerYes{}
  \item Did you describe the limitations of your work?
    \answerYes{Both at the end of the Section~\ref{sec:numerics} and the future work discussed in Section~\ref{sec:conclude}.}
  \item Did you discuss any potential negative societal impacts of your work?
    \answerNA{}
  \item Have you read the ethics review guidelines and ensured that your paper conforms to them?
    \answerYes{}
\end{enumerate}

\item If you are including theoretical results...
\begin{enumerate}
  \item Did you state the full set of assumptions of all theoretical results?
    \answerNA{}
	\item Did you include complete proofs of all theoretical results?
    \answerNA{}
\end{enumerate}

\item If you ran experiments...
\begin{enumerate}
  \item Did you include the code, data, and instructions needed to reproduce the main experimental results (either in the supplemental material or as a URL)?
    \answerYes{The code is included as part of a software package, available at: \url{https://github.com/ywa136/TensorGraphicalModels.jl}}
  \item Did you specify all the training details (e.g., data splits, hyperparameters, how they were chosen)?
    \answerYes{}
	\item Did you report error bars (e.g., with respect to the random seed after running experiments multiple times)?
    \answerYes{}
	\item Did you include the total amount of compute and the type of resources used (e.g., type of GPUs, internal cluster, or cloud provider)?
    \answerYes{All numerical experiments were run on a laptop equipped with \texttt{Intel Core i5 Quad-Core 1.4 GHz} with a \texttt{16 GB RAM}.}
\end{enumerate}

\item If you are using existing assets (e.g., code, data, models) or curating/releasing new assets...
\begin{enumerate}
  \item If your work uses existing assets, did you cite the creators?
    \answerNA{}
  \item Did you mention the license of the assets?
    \answerNA{}
  \item Did you include any new assets either in the supplemental material or as a URL?
    \answerNA{}
  \item Did you discuss whether and how consent was obtained from people whose data you're using/curating?
    \answerNA{}
  \item Did you discuss whether the data you are using/curating contains personally identifiable information or offensive content?
    \answerNA{}
\end{enumerate}

\item If you used crowdsourcing or conducted research with human subjects...
\begin{enumerate}
  \item Did you include the full text of instructions given to participants and screenshots, if applicable?
    \answerNA{}
  \item Did you describe any potential participant risks, with links to Institutional Review Board (IRB) approvals, if applicable?
    \answerNA{}
  \item Did you include the estimated hourly wage paid to participants and the total amount spent on participant compensation?
    \answerNA{}
\end{enumerate}

\end{enumerate}

\end{document}